\documentclass[conference]{IEEEtran}
\IEEEoverridecommandlockouts
\usepackage{cite}
\usepackage{amsmath,amssymb,amsfonts}
\usepackage{algorithmic}
\usepackage{graphicx}
\usepackage{textcomp}
\usepackage{xcolor}
\usepackage{verbatim}
\usepackage{multirow}
\usepackage{color,xcolor,soul}
\usepackage{fancyhdr}
\def\BibTeX{{\rm B\kern-.05em{\sc i\kern-.025em b}\kern-.08em
    T\kern-.1667em\lower.7ex\hbox{E}\kern-.125emX}}

\begin{document}

\title{Hyperparameter Optimization in Binary Communication Networks for Neuromorphic Deployment\\
\thanks{Notice: This manuscript has been authored in part by UT-Battelle, LLC under Contract No. DE-AC05-00OR22725 with the U.S. Department of Energy. The United States Government retains and the publisher, by accepting the article for publication, acknowledges that the United States Government retains a non-exclusive, paid-up, irrevocable, world-wide license to publish or reproduce the published form of this manuscript, or allow others to do so, for United States Government purposes. The Department of Energy will provide public access to these results of federally sponsored research in accordance with the DOE Public Access Plan (http://energy.gov/downloads/doe-public-access-plan).}
}

\makeatletter
    \newcommand{\linebreakand}{%
      \end{@IEEEauthorhalign}
      \hfill\mbox{}\par
      \mbox{}\hfill\begin{@IEEEauthorhalign}
    }
    \makeatother

\author{\IEEEauthorblockN{Maryam Parsa}
\IEEEauthorblockA{\textit{Purdue University} \\
West Lafayette, Indiana, USA \\
mparsa@purdue.edu}
\and
\IEEEauthorblockN{Catherine D. Schuman}
\IEEEauthorblockA{\textit{Oak Ridge National Laboratory}\\
Oak Ridge, Tennessee, USA \\
schumancd@ornl.gov}
\and
\IEEEauthorblockN{Prasanna Date}
\IEEEauthorblockA{\textit{Oak Ridge National Laboratory} \\
Oak Ridge, Tennessee, USA\\
datepa@ornl.gov}
\linebreakand
\IEEEauthorblockN{Derek C. Rose}
\IEEEauthorblockA{\textit{Oak Ridge National Laboratory} \\
Oak Ridge, Tennessee, USA\\
rosedc@ornl.gov}
\and
\IEEEauthorblockN{Bill Kay}
\IEEEauthorblockA{\textit{Oak Ridge National Laboratory} \\
Oak Ridge, Tennessee, USA\\
kaybw@ornl.gov}
\and
\IEEEauthorblockN{J. Parker Mitchell}
\IEEEauthorblockA{\textit{Oak Ridge National Laboratory} \\
Oak Ridge, Tennessee, USA\\
mitchelljp1@ornl.gov}
\linebreakand
\IEEEauthorblockN{Steven R. Young}
\IEEEauthorblockA{\textit{Oak Ridge National Laboratory} \\
Oak Ridge, Tennessee, USA\\
youngsr@ornl.gov}
\and
\IEEEauthorblockN{Ryan Dellana}
\IEEEauthorblockA{\textit{Sandia National Laboratories} \\
Albequerque, New Mexico, USA \\
rdellan@sandia.gov}
\and
\IEEEauthorblockN{William Severa}
\IEEEauthorblockA{\textit{Sandia National Laboratories} \\
Albequerque, New Mexico, USA \\
wmsever@sandia.gov}
\linebreakand
\IEEEauthorblockN{Thomas E. Potok}
\IEEEauthorblockA{\textit{Oak Ridge National Laboratory} \\
Oak Ridge, Tennessee, USA\\
potokte@ornl.gov}
\and
\IEEEauthorblockN{Kaushik Roy}
\IEEEauthorblockA{\textit{Purdue University} \\
West Lafayette, Indiana, USA \\
kaushik@purdue.edu}
}

\maketitle
\thispagestyle{fancy}
\begin{abstract}
Training neural networks for neuromorphic deployment is non-trivial. There have been a variety of approaches proposed to adapt back-propagation or back-propagation-like algorithms appropriate for training. Considering that these networks often have very different performance characteristics than traditional neural networks, it is often unclear how to set either the network topology or the hyperparameters to achieve optimal performance.  In this work, we introduce a Bayesian approach for optimizing the hyperparameters of an algorithm for training binary communication networks that can be deployed to neuromorphic hardware. We show that by optimizing the hyperparameters on this algorithm for each dataset, we can achieve improvements in accuracy over the previous state-of-the-art for this algorithm on each dataset (by up to 15 percent). This jump in performance continues to emphasize the potential when converting traditional neural networks to binary communication applicable to neuromorphic hardware.

\end{abstract}


\section{Introduction}


Neuromorphic computing offers the promise of very low power hardware implementations of machine learning, along with potential opportunities for new ways to perform computing with a fundamentally different type of architecture \cite{aimone2018non}. Neuromorphic hardware platforms are being developed by both industry and academic groups that have largely focused on providing an implementation of traditional spiking neural networks. To date there has been relatively little focus on the development of algorithms which aim to effectively leverage neuromorphic systems for spiking networks \cite{schuman2017survey}.

One common class of algorithms of this type are based on traditional back-propagation-trained algorithms, such as those used for traditional neural network training but have been adapted to accommodate for neuromorphic deployment. When these algorithms are applied to networks that can be deployed on spiking neuromorphic systems, hyperparameters can have a tremendous impact on the performance of the network.  This challenge is well known in traditional neural network training \cite{bergstra2013hyperopt, hernandez2016general, parsa2019pabo}, and often these hyperparameters are determined through a combination of trial-and-error, intuition, and random search \cite{bergstra2012random}.  However, it is not clear how these methods should be adapted to accommodate for the changes in the training algorithm.  Moreover, there are often even more new hyperparameters for these adapted approaches to back-propagation-like algorithms.

One such algorithm we choose to investigate is Whetstone \cite{severa2019training}.  Whetstone trains networks that have binary communication, which are amenable for mapping onto spiking neuromorphic hardware.  In this approach, neural networks are trained initially with differentiable activation functions (e.g., sigmoidal or bounded rectified linear units), but over the course of gradient descent optimization, the activation functions are slowly ``sharpened" to non-differentiable threshold functions.  This approach not only has all of the hyperparameters associated with traditional neural network or deep learning network training, but also additional hyperparameters of its own, for example, associated with how sharpening occurs over the course of the algorithm.  As we will show below, these hyperparameters can have a significant effect on the performance of the algorithm, but it is not clear what hyperparameters to use for a given dataset \textit{a priori}.

In this work, we apply Bayesian hyperparameter optimization~\cite{parsa2019pabo, parsa2019bayesian, parsa2020journal} to find optimal hyperparameters for the Whetstone algorithm on four different datasets. We compare our results to the previously published Whetstone results from \cite{severa2019training} and show that by tuning the hyperparameters for each dataset we can achieve significantly better performance, up to a $15\%$ improvement in accuracy in some cases. We compare the best performing hyperparameters for each dataset, and study the sensitivity of the final performance on the changes of hyperparameters. Finally, we discuss how this approach can be extended, both in future work with the Whetstone approach as well as other training approaches for Spiking Neural Networks (SNNs), and neuromorphic systems. These results represent, not just an improvement over state-of-the-art, but also an indication that off-the-shelf spiking algorithms may be significantly improved by optimization via this Bayesian approach.

The main contributions of this work are:
\begin{itemize}
    \item A demonstration of the effect of hyperparameters on a training algorithm (Whetstone) that trains neural networks with binary communication.
    \item A Bayesian optimization approach to optimize Whetstone's hyperparameters.
    \item State-of-the-art results for Whetstone on four commonly used datasets. 
\end{itemize}

\section{Background and Related Work}
\label{background}

We first review the various approaches used for optimizing the hyperparameters of deep learning models. Hyperparameter optimization for neural networks used to be largely governed by rules of thumb \cite{date2016design}.
Bengio outlines some of these rules and practical guidelines for efficiently training large-scale deep neural networks \cite{bengio2012practical}. Bergstra and Bengio show that random search outperforms grid search and manual search for hyperparameter optimization and has good theoretical guarantees and empirical evidence \cite{bergstra2012random}.
Continuing along this line of research, Bergstra \emph{et al.} present greedy sequential algorithms for hyperparameter optimization and show that their performance is better than that of random search \cite{bergstra2011algorithms}.

Bayesian-based approaches have also been used for optimizing the hyperparameters of deep neural networks. Bergstra \emph{et al.} show that algorithms based specifically on the Gaussian process are the most call-efficient for hyperparameter optimization of deep neural networks \cite{bergstra2014preliminary}.
Snoek \emph{et al.} describe algorithms that take into consideration the variable costs of learning experiments and show that the resulting set of hyperparameters returned by these algorithms can match or even surpass human expert-level optimizations \cite{snoek2012practical}.
Zhang \emph{et al.} propose a search algorithm based on Bayesian optimization while training deep convolutional neural networks on the PASCAL VOC 2007 and 2012 datasets \cite{zhang2015improving}.
Balaprakash \emph{et al.} develop DeepHyper, which is a Python package that leverages the Balsam workflow and provides an interface for implementation and study of scalable hyperparameter search methods \cite{balaprakash2018deephyper}.
Ilievski \emph{et al.} propose a deterministic and efficient method for hyperparameter optimization using radial basis function as the error surrogate in Bayesian-based methods called HORD, and demonstrate its effectiveness on MNIST and CIFAR-10 datasets \cite{ilievski2017efficient}. 



Evolutionary optimization techniques have also been used for hyperparameter optimization in the literature.
Miikkulainen \emph{et al.} propose CoDeepNEAT, which is a method that extends the conventional neuro-evolution methods to topology, components and hyperparameters and achieves performance comparable to the best human-optimized networks \cite{miikkulainen2019evolving}.
Young \emph{et al.} propose a scalable evolutionary optimization method and demonstrate its efficacy on varied datasets \cite{young2017evolving}.
Shafiee \emph{et al.} propose a genetic algorithm-like method for hyperparameter optimization, which not only achieves the state-of-the-art performance, but is also seen to use up to $48 \times$ less synapses in doing so \cite{shafiee2018deep}.
Liang \emph{et al.} evaluate several hyperparameter optimization methods that evolve the architecture of deep neural networks and demonstrate that a synergetic approach for evolving custom routings with evolved, shared modules is very powerful, and significantly improves the state-of-the-art performance on the Omniglot character recognition domain \cite{liang2018evolutionary}.
In addition to these evolutionary optimization-based methods, reinforcement learning has also been used for hyperparameter optimization of deep neural networks.



While the above approaches catered to deep neural networks, several hyperparameter optimization methods have been used in the literature for optimizing architectures or hyperparameters specifically pertaining to neuromorphic computing.
Schuman \emph{et al.} present several approaches for encoding numerical values as spikes for spiking neural networks, hierarchically combine them to form more complex encoding schemes, and demonstrate their usability on four different applications \cite{schuman2019non}.
Salt \emph{et al.} use differential evolution (DE) and self-adaptive differential evolution algorithms (SADE) to optimize the parameter space of synaptic plasticity and membrane adaptivity learning mechanisms in the lobula giant movement detector (LGMD) neuron that is driven by a dynamic vision sensor (DVS) camera \cite{salt2017differential}. 
Schuman \emph{et al.} develop an evolutionary optimization based training framework for spiking neural network and neuromorphic architectures, and test this approach on four datasets \cite{schuman2016evolutionary}.
Kim and Kim apply a Neuro-evolutionary algorithm to optimize the hyperparameters of spiking neural networks and show that the model trained using this approach outperforms all other models \cite{kim2018competitive}.
Parsa \emph{et al.} demonstrate effectiveness of Bayesian approach for hyperparameter optimization for spiking neuromorphic systems \cite{parsa2019bayesian}.

In this work we focus on Bayesian hyperparameter optimization for binary communication network for neuromorphic deployment, Whetstone~\cite{severa2019training}. The powerful and yet effective underlying mathematics of Bayesian approach, paves the way to quickly estimate an expensive objective function such as network performance.








\section{Methods}
In this section we briefly introduce Whetstone and Bayesian optimization approaches. The former is an approach for training binary communication networks for neuromorphic deployment, and the latter is an optimization tool for problems with black-box and expensive objective functions. Detailed description on each of these techniques can be found at \cite{severa2019training}, and \cite{shahriari2015taking, parsa2020journal}, respectively.

\subsection{Whetstone}
Whetstone utilizes bounded rectified linear units (bRELUs) and sigmoidal units that are modified during training to approach binarized step-functions. The approach aims to gradually modify the activation function so as to minimally otherwise disrupt network training. Due to the sensitivity of backpropagation to zeroed activations, this sharpening and thus binary conversion process was found to be more stable when applied layer-by-layer on a schedule and in the direction of input layer to output layer. This \textit{scheduled-sharpening} involves several hyperparameters, such as the epoch to start the sharpening, duration of sharpening, and number of epochs to wait before starting the next scheduled sharpening (intermission). To avoid a fully manual schedule with additional hyperparameters, Whetstone's authors introduced an \textit{adaptive-sharpening} scheduler that monitors loss after each training epoch and decides to resume or pause sharpening dependent on the relative change in training loss.

Whetstone also attempts to mitigate a condition which occurs in bRELUs and sigmoidal nodes that stop responding and produce zero outputs regardless of input.  The authors note that this condition happens in non-binarized networks as well but hypothesize that the sharpening process can increase occurrence odds.  To alleviate this problem, Whetstone networks typically use redundant output encodings as output targets. To produce output for loss computation, Whetstone uses a softmax over a population encoding (neuron distribution key generated or specified at network initialization) that allows for $n$-hot encoding of targets while output neurons can contribute to more than one class.  This also enables the use of a cross-entropy loss function (common to many neural network classification tasks), which the authors found to be more effective than a direct mean squared error vector loss.



Severa \textit{et al.} \cite{severa2019training} also demonstrate the effects of architecture hyperparameters such as number of convolution layers and filter sizes on the overall performance of Whetstone for four different dataset. Their results for these various hyperparameters were consistent with the intuition that deeper networks perform better for spiking networks. However, they did not perform any comprehensive hyperparameter optimization. Additional instability was noted in relation to the choice of optimizer used during training, with Adam optimized networks' performance being especially sensitive to initial conditions. For the choice of optimizer, they show that Adadelta and RMSprop are more reliable compared to Adam.  Batch normalization was further found to improve stability during training. The sensitivity of Whetstone approach on various hyperparameters such as the choice of optimizer or batch normalization layer, differentiates the hyperparameter optimization approach for this binary communication from traditional artificial neural network training. This leads to a research question on which hyperparameter optimization technique is suitable for non-traditional networks such as Whetstone.

In this work, we only focus on \textit{scheduled-sharpening} due to the stability and consistency of the results obtained with this scheduler. In our hyperparameter optimization search, we considered three main hyperparameters involved in this technique: sharpener starting epoch (``sh\_st"), duration (``sh\_du"), and intermission (``sh\_int"). For each case study, detailed of the ranges for each of these hyperparameters is given in the following section.

\subsection{Bayesian Optimization}

To systematically take the human out of the loop in finding the optimum set of hyperparameter for an expensive, black-box objective function such as training a neural networks, several approaches are introduced in the literature and already discussed in section~\ref{background}. Bayesian optimization is one of the primary approaches for these types of problems due to its flexible and powerful underlying mathematics~\cite{shahriari2015taking}.

As summarized by~\cite{shahriari2015taking, parsa2019pabo, parsa2020journal}, Bayesian optimization is a sequential technique that aims at predicting the unknown objective function with limited and yet effective observations. For our hyperparameter optimization problem, the unknown objective function is the classification performance of Whetstone, and observations are the performance values (accuracies) for a set of hyperparameters in each iteration. We start the optimization process with two random initial set of hyperparameters, and for each one of them evaluate the performance of Whetstone network. This will create the first set of observations. In the Bayesian optimization technique for each iteration, we estimate a Gaussian distribution over the available observations (called the \textit{prior} distribution, current beliefs). We update the current beliefs with a new observation and estimate the \textit{posterior} distribution. With enough observations, the \textit{posterior} distribution is the prediction of the unknown, expensive objective function we are optimizing. In this search technique, the new observations are chosen based on optimizing a surrogate model, called the \textit{acquisition function}. This function is built upon the \textit{posterior} distribution at each iteration. There are different policies introduced in the literature to calculate this function such as improved-based, optimistic, and information-based policies. Each one of these approaches calculate the \textit{acquisition function} to explore and exploit the search space. The maximum point of this function is the best next set of hyperparameter to observe in the next iteration. More details on Bayesian optimization can be found in \cite{shahriari2015taking}.  In this work, we are dealing with a single objective Bayesian optimization problem~\cite{skopt:2020}, as we aim at finding the optimum set of hyperparameter that maximizes the Whetstone performance.

\section{Results}

    
    
    
    


We validate our Bayesian hyperparameter optimization approach across several datasets, hyperparameter combinations and case studies. In using Whetstone, there are a variety of sets of hyperparameters that can be optimized.  Here we focus on the following hyperparameter sets: optimizer parameters, noise parameters, batch normalization parameters, Whetstone sharpener parameters, and CNN architecture parameters. Details of the hyperparameters that are optimized and their corresponding ranges are given in each case study as follows. 

\subsection{Datasets}



Our methods were benchmarked on four labeled image data sets commonly used to demonstrate efficacy of supervised image classification protocols. The {\em MNIST~\cite{lecun2010mnist}} dataset consists of  gray-scale images of handwritten single digits, each $28 \times 28$ pixels. There are 10 classes, one for each number $0-9$, and the data is split in to a training set of $60000$ images and a test set of $10000$ images. The {\em Fashion MNIST~\cite{xiao2017fashion}}  dataset consists of gray-scale images of miscellaneous clothing  items (shirts, pants, shoes, etc.), each $28 \times 28$ pixels. There are 10 classes, one for each type of item, and the data is split in to a training set of $60000$ images and a test set of $10000$ images. The Fashion MNIST dataset is designed to be a drop in replacement for the MNIST dataset, with the only difference being the items which are classified. The {\em CIFAR-10~\cite{krizhevsky2009learning}} dataset consists of  color images of miscellaneous items (dogs, airplanes, birds, ships, etc.), each $32 \times 32$ pixels. There are 10 classes, one for each type of item, and the data is split in to a training set of $50000$ images and a test set of $10000$ images. The {\em CIFAR-100}~\cite{krizhevsky2009learning} dataset is the same as the CIFAR-10 dataset, except with 100 classes. Each class represents an equal proportion of the total dataset. 

\subsection{Case Study One}

For case study one, we select a small search space for hyperparameters given in Table~\ref{tab:cs1_hp} for classification task on CIFAR-100 dataset~\cite{krizhevsky2009learning}. This limited search space is helpful in validating the results through comparing the optimum hyperparameters from the optimization technique and the grid search approach. The grid search approach is evaluating the network for all possible combinations of the hyperparameters. In this case study, the fixed hyperparameters that are not included in the optimization search and their corresponding values are given in Table~\ref{tab:cs1_hp_fixed}.

\begin{table}[]
\centering
\caption{Case study one: Evaluated hyperparameters}
\label{tab:cs1_hp}
\begin{tabular}{|l|l|}
\hline
\textbf{Hyperparameter} & \textbf{Range} \\ \hline \hline
Optimizer learning rate (lr) & 0.0001, 1 \\ \hline
Optimizer decay (dec) & 1e-8, 1e-6 \\ \hline
Sharpener starting epoch (sh\_st) & 15, 25 \\ \hline
Sharpener duration (sh\_du) & 3, 7 \\ \hline
Sharpener intermission (sh\_int) & 2, 5 \\ \hline
Conv. layer 1, filter size (filter 1) & 3, 7 \\ \hline
Conv. layer 1, \# of features (feat 1) & 64, 128 \\ \hline
Dense layer, \# of features (dense) & 256, 1024 \\ \hline 
\multicolumn{2}{|c|}{\textbf{Search space size: 256}} \\ \hline
\end{tabular}
\end{table}

\begin{table}[]
\centering
\vspace{-0.2cm}
\label{tab:cs1_hp_fixed}
\begin{tabular}{|l|l|}
\hline
\textbf{Hyperparameter} & \textbf{Value} \\ \hline \hline
Optimizer rho & 0.9 \\ \hline
Optimizer epsilon & 1e-6 \\ \hline
Optimizer type & Adadelta \\ \hline
Gaussian noise layer & Without noise \\ \hline
\begin{tabular}[c]{@{}l@{}}Batch normalizer\\ conv. layers momentum\end{tabular} & 0.95 \\ \hline
\begin{tabular}[c]{@{}l@{}}Batch normalizer\\ dense layer momentum\end{tabular} & 0.95 \\ \hline
Batch normalizer epsilon & 1e-3 \\ \hline
Batch normalizer center & True \\ \hline
Batch normalizer scale & True \\ \hline
Conv. layer 2, filter size & 5 \\ \hline
Conv. layer 3, filter size & 3 \\ \hline
Conv. layer 2, \# of features & 256 \\ \hline
Conv. layer 3, \# of features & 512 \\ \hline
\end{tabular}
\end{table}

\begin{figure}
    \centering
    \includegraphics[scale=.4]{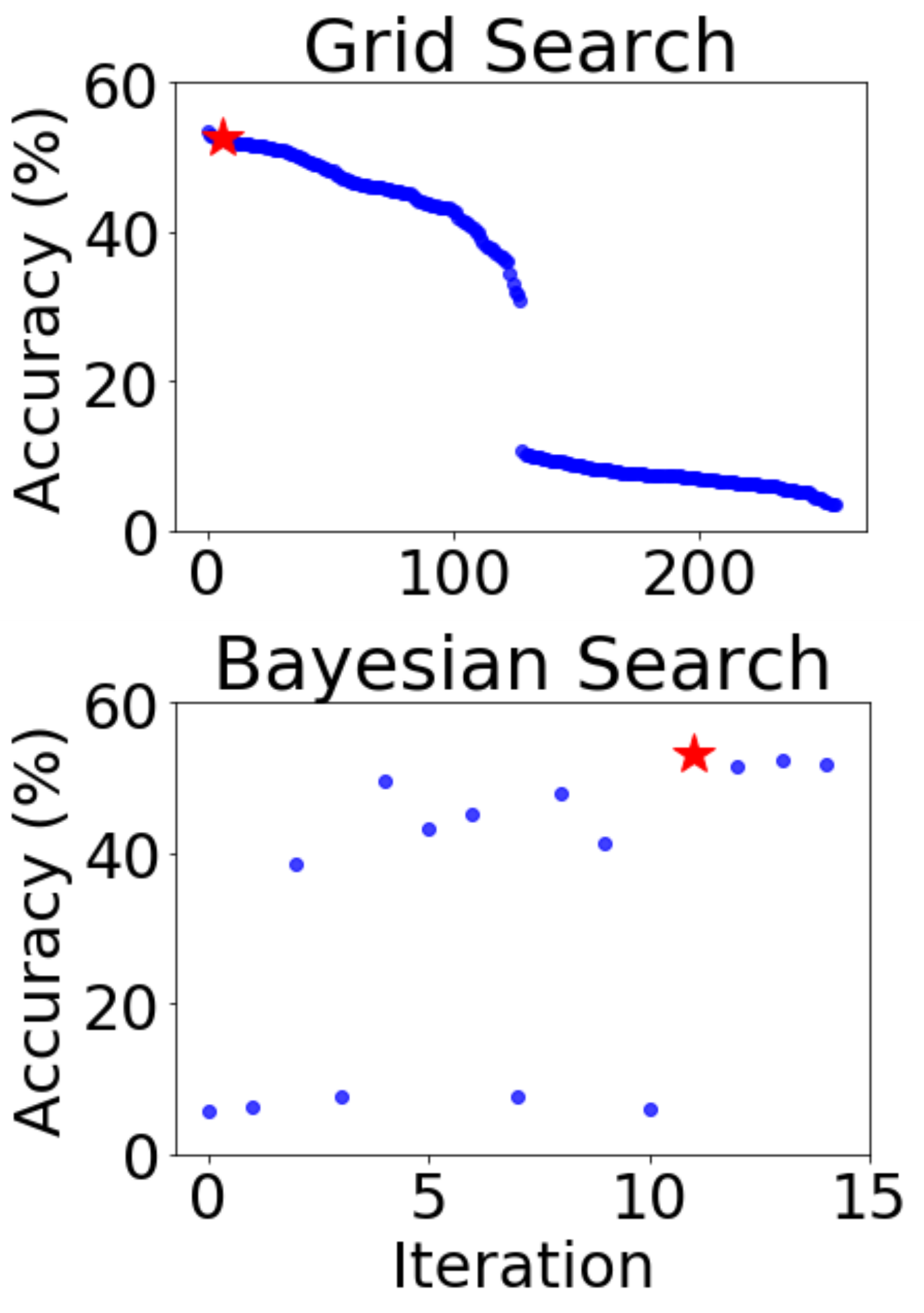}
    \vspace{-0.2cm}
    \caption{Case Study One: Comparing grid search and Bayesian hyperparameter optimization for hyperparameters given in Table~\ref{tab:cs1_hp} with search space size of 256}
    \label{fig:cs_1}
\end{figure}

\begin{table*}[]
\centering
\caption{Case study one: Details of Bayesian hyperparameter optimization}
\label{tab:cs1_results}
\begin{tabular}{|l|c|c|c|c|c|c|c|c|c|c|c|c|c|c|c|}
\hline
HPs & Iter 1 & Iter 2 & Iter 3 & Iter 4 & Iter 5 & Iter 6 & Iter 7 & Iter 8 & Iter 9 & Iter 10 & Iter 11 & \textbf{Iter 12} & Iter 13 & Iter 14 & Iter 15 \\ \hline \hline
lr & 1e-4 & 1e-4 & 1 & 1e-4 & 1 & 1 & 1 & 1e-4 & 1 & 1 & 1e-4 & \textbf{1} & 1 & 1 & 1 \\ 
dec & 1e-8 & 1e-6 & 1e-6 & 1e-8 & 1e-6 & 1e-6 & 1e-6 & 1e-6 & 1e-6 & 1e-6 & 1e-6 & \textbf{1e-6} & 1e-8 & 1e-6 & 1e-6 \\ 
sh\_st & 25 & 25 & 15 & 15 & 25 & 25 & 25 & 25 & 25 & 25 & 25 & \textbf{25} & 25 & 25 & 25 \\ 
sh\_dur & 3 & 7 & 3 & 3 & 7 & 7 & 3 & 3 & 7 & 3 & 7 & \textbf{7} & 7 & 7 & 7 \\ 
sh\_int & 2 & 2 & 2 & 5 & 5 & 5 & 5 & 2 & 2 & 5 & 2 & \textbf{2} & 5 & 5 & 2 \\ 
filter 1 & 3 & 7 & 7 & 3 & 3 & 7 & 7 & 3 & 7 & 7 & 3 & \textbf{3} & 3 & 3 & 3 \\ 
feat 1 & 64 & 128 & 128 & 64 & 64 & 64 & 64 & 128 & 128 & 64 & 128 & \textbf{128} & 64 & 128 & 64 \\ 
dense & 256 & 256 & 1024 & 1024 & 256 & 256 & 1024 & 1024 & 1024 & 256 & 256 & \textbf{1024} & 1024 & 1024 & 1024 \\ \hline
\textbf{Acc (\%)} & \textbf{5.61} & \textbf{6.37} & \textbf{38.65} & \textbf{7.69} & \textbf{49.69} & \textbf{43.4} & \textbf{45.19} & \textbf{7.59} & \textbf{47.84} & \textbf{41.34} & \textbf{6.11} & \textbf{53.13} & \textbf{51.54} & \textbf{52.38} & \textbf{51.69} \\ \hline
\end{tabular}
\end{table*}

In Figure~\ref{fig:cs_1}, for CIFAR-100 dataset, the grid search results are compared with the results from the Bayesian hyperparameter search. The hyperparameter ranges are given in Table~\ref{tab:cs1_hp}. After only 15 evaluations of Whetstone~\cite{severa2019training}, the Bayesian hyperparameter search finds the almost optimum combination of hyperparameters that the grid search predicts after 256 evaluations. This optimal point for the Bayesian search, $(l\_r = 1, dec = 1e-6, sh\_st = 25, sh\_du = 7, sh\_int = 2, filter 1 = 3, feat 1 = 128, dense = 1024)$, is shown in red star in Figure~\ref{fig:cs_1}, and leads to accuracy of $53.13\%$, which outperforms the $38\%$ accuracy reported in Whetstone original results~\cite{severa2019training}. The optimum hyperparameter set for the grid search is $(l\_r = 1, dec = 1e-6, sh\_st = 25, sh\_du = 3, sh\_int = 5, filter 1 = 3, feat 1 = 128, dense = 1024)$ with accuracy of $53.34\%$. These two points predict almost the same classification accuracy and only differ in two hyperparameters of ``duration of sharpening", and ``sharpening intermission". The hyperparameter values at each iteration are given in Table~\ref{tab:cs1_results}.

We also perform further analysis on the changes of hyperparameters and their effect on the final accuracy of the network. For example, with changing the sharpener starting epoch from 15 to 25, its duration from 3 to 7, and the filter size in the first convolution layer from 7 to 3, we are able to improve the final accuracy from $38.65\%$ to $53.13\%$ (iteration 3 versus iteration 13 in Table~\ref{tab:cs1_results}). 
This table also shows that some hyperparameters play a vital role on the final performance of the system, such as learning rate.

\vspace{-0.2cm}
\subsection{Case Study Two}

In case study two, we increase the search space size to 398,131,200 combinations of hyperparameters shown in Table~\ref{tab:cs2_hp}. In this scenario we consider various hyperparameter types ranging from optimizer hyperparameters, to Gaussian noise, or batch normalization layers. In addition we also include the Whetstone scheduled sharpening ~\cite{severa2019training} hyperparameters, and the hyperparameters that belong to the neural network architecture itself, such as filter sizes or the number of features to extract.  

The Whetstone's scheduled sharpener sharpens layers one at a time in sequential order. The ``start epoch" hyperparameter is the epoch on which it begins sharpening the first layer. The ``duration" is how many epochs it takes to sharpen each layer, and the ``intermission" is how many epochs it waits after sharpening a layer before beginning sharpening of the next layer. For each hyperparameter, all values in Table~\ref{tab:cs2_hp} are based on acceptable and reasonable ranges.

\begin{table}[]
\centering
\caption{Case study two: Evaluated Hyperparameters}

\label{tab:cs2_hp}
\begin{tabular}{|l|l|l|}
\hline
 & \textbf{Hyperparameter} & \textbf{Options} \\ \hline \hline
\multirow{5}{*}{Optimizer} & Learning rate & \begin{tabular}[c]{@{}l@{}} 0.0001, 0.001, 0.01, \\ 0.1, 1 \end{tabular} \\ \cline{2-3} 
 & Rho & 0.9, 0.95 \\ \cline{2-3} 
 & Epsilon & 1e-8, 1e-6 \\ \cline{2-3} 
 & Decay & 1e-8, 1e-6 \\ \cline{2-3} 
 & Type & Adadelta, RMSprop \\ \hline
\multirow{2}{*}{Noise} & Standard deviation & 0.2, 0.3 \\ \cline{2-3} 
 & Location & \begin{tabular}[c]{@{}l@{}}Without noise, \\ After first dense layer\end{tabular} \\ \hline
\multirow{5}{*}{\begin{tabular}[c]{@{}l@{}}Batch \\ Normalizer\end{tabular}} & Momentum, conv. & 0.85, 0.95 \\ \cline{2-3} 
 & Momentum, dense & 0.85, 0.95 \\ \cline{2-3} 
 & Epsilon & 1e-3, 1e-2 \\ \cline{2-3} 
 & Center & False, True \\ \cline{2-3} 
 & Scale & False, True \\ \hline
\multirow{3}{*}{\begin{tabular}[c]{@{}l@{}}Sharpener \\ Schedule\end{tabular}} & Start Epoch & 20, 25, 30 \\ \cline{2-3} 
 & Duration & 4, 5, 6, 7 \\ \cline{2-3} 
 & Intermission & 1, 2, 3, 4, 5 \\ \hline
\multirow{7}{*}{\begin{tabular}[c]{@{}l@{}}CNN\\ Architecture\end{tabular}} & Conv. layer 1, filter size & 3, 5, 7 \\ \cline{2-3} 
 & Conv. layer 2, filter size & 3, 5 \\ \cline{2-3} 
 & Conv. layer 3, filter size & 3, 5 \\ \cline{2-3} 
 & Conv. layer 1, \# of features & 32, 64, 128 \\ \cline{2-3} 
 & Conv. layer 2, \# of features & 64, 128, 256 \\ \cline{2-3} 
 & Conv. layer 3, \# of features & 256, 512 \\ \cline{2-3} 
 & Dense layer, \# of features & 256, 512, 1024 \\ \hline
 
 \multicolumn{3}{|c|}{\textbf{Search Space Size: 398,131,200}} \\ \hline
\end{tabular}
\end{table}

For the hyperparameters given in Table~\ref{tab:cs2_hp}, the performance of the hyperparameter optimization approach for Whetstone technique for four different dataset of MNIST~\cite{lecun2010mnist}, Fashion-MNIST~\cite{xiao2017fashion}, CIFAR-10~\cite{krizhevsky2009learning}, and CIFAR-100~\cite{krizhevsky2009learning} as well as their corresponding optimum hyperparameter values are given in Table~\ref{tab:cs2_acc}. For each dataset, the Whetstone network is trained for 50 epochs and the hyperparameter optimization search evaluated the network for 30 different hyperparameter sets. The Whetstone performance once its hyperparameters are optimized is increased from $99.53\%$ to $99.6\%$ for MNIST, and from $93.2\%$ to $93.68\%$ for Fashion-MNIST dataset. This improved performance is more noticeable for larger dastaset such as CIFAR-10 and CIFAR-100. For the former, the accuracy is increased from $79\%$ to $84.36$, and for the latter it is improved from $38\%$ to $53.42\%$. Table~\ref{tab:literature} shows a comparison between the Spiking Neural Network (SNN) classification accuracies on MNIST, Fashion-MNIST, CIFAR-10, and CIFAR-100 dataset for state-of-the-art models and network architectures in the literature. The purpose of this work is not obtaining the best accuracy for each dataset; instead, our goal is to show that with an effective hyperparameter optimization framework, we can drastically improve a performance of a model with only few evaluations. It is worth noting that the networks that achieve higher accuracy in this table are often significantly more complicated than the network structure we use, in terms of the architecture and input encoding techniques for SNNs.  By allowing for more complex network structures, we expect that comparable accuracies can be achieved. 

\begin{table*}[] \centering
\caption{Case study two: Optimized hyperparameters and their corresponding classification accuracies for different dataset}
\label{tab:cs2_acc}
\begin{tabular}{l|l|c|c|c|c}
\multicolumn{2}{c|}{Dataset}  & MNIST & Fashion-MNIST & CIFAR-10 & CIFAR-100 \\ \hline \hline
\multirow{5}{*}{\begin{tabular}[c]{@{}l@{}} Optimizer Hyperparameters\end{tabular}} & Learning Rate & 0.001 & 0.001 & 0.001 & 1 \\
 & Rho & 0.95 & 0.9 & 0.9 & 0.9 \\
 & Epsilon & 1e-6 & 1e-6 & 1e-8 & 1e-6 \\
 & Decay & 1e-8 & 1e-6 & 1e-6 & 1e-6 \\
 & Type & RMSprop & RMSprop & RMSprop & Adadelta \\ \hline
\multirow{2}{*}{\begin{tabular}[c]{@{}l@{}}Noise Layer\\ Hyperparameters\end{tabular}} & Standard deviation & - & 0.2 & - & - \\
 & Location & No Noise & After 1st Dense & No Noise & No Noise \\ \hline
\multirow{5}{*}{\begin{tabular}[c]{@{}l@{}}Batch Normalizer\\ Hyperparameters\end{tabular}} & Momentum, conv. & 0.95 & 0.95 & 0.85 & 0.95 \\
 & Momentum, dense & 0.95 & 0.85 & 0.95 & 0.95 \\
 & Epsilon & 1e-2 & 1e-2 & 1e-3 & 1e-3 \\
 & Center & False & False & True & False \\
 & Scale & False & False & False & False \\ \hline
\multirow{3}{*}{\begin{tabular}[c]{@{}l@{}}Whetstone Sharpener Schedule\\  Hyperparameters\end{tabular}} & Start Epoch & 30 & 20 & 30 & 30 \\
 & Duration & 6 & 4 & 4 & 4 \\
 & Intermission & 4 & 5 & 2 & 5 \\ \hline
\multirow{7}{*}{\begin{tabular}[c]{@{}l@{}}CNN Architecture\\ Hyperparameters\end{tabular}} & Conv. layer 1, filter size & 7 & 3 & 3 & 3 \\
 & Conv. layer 2, filter size & 5 & 3 & 5 & 5 \\
 & Conv. layer 3, filter size & 3 & 5 & 5 & 5 \\
 & Conv. layer 1, \# of features & 128 & 128 & 64 & 128 \\
 & Conv. layer 2, \# of features & 128 & 128 & 256 & 256 \\
 & Conv. layer 3, \# of features & 256 & 512 & 512 & 512 \\
 & Dense layer, \# of features & 256 & 512 & 512 & 1024 \\ \hline \hline
\multicolumn{2}{c|}{\textbf{Accuracy}} & \textbf{99.6\%} & \textbf{93.68\%} & \textbf{83\%} & \textbf{53.42\%} \\ 
\end{tabular}
\end{table*}

\begin{figure}
    \centering
    \includegraphics [scale=.25]{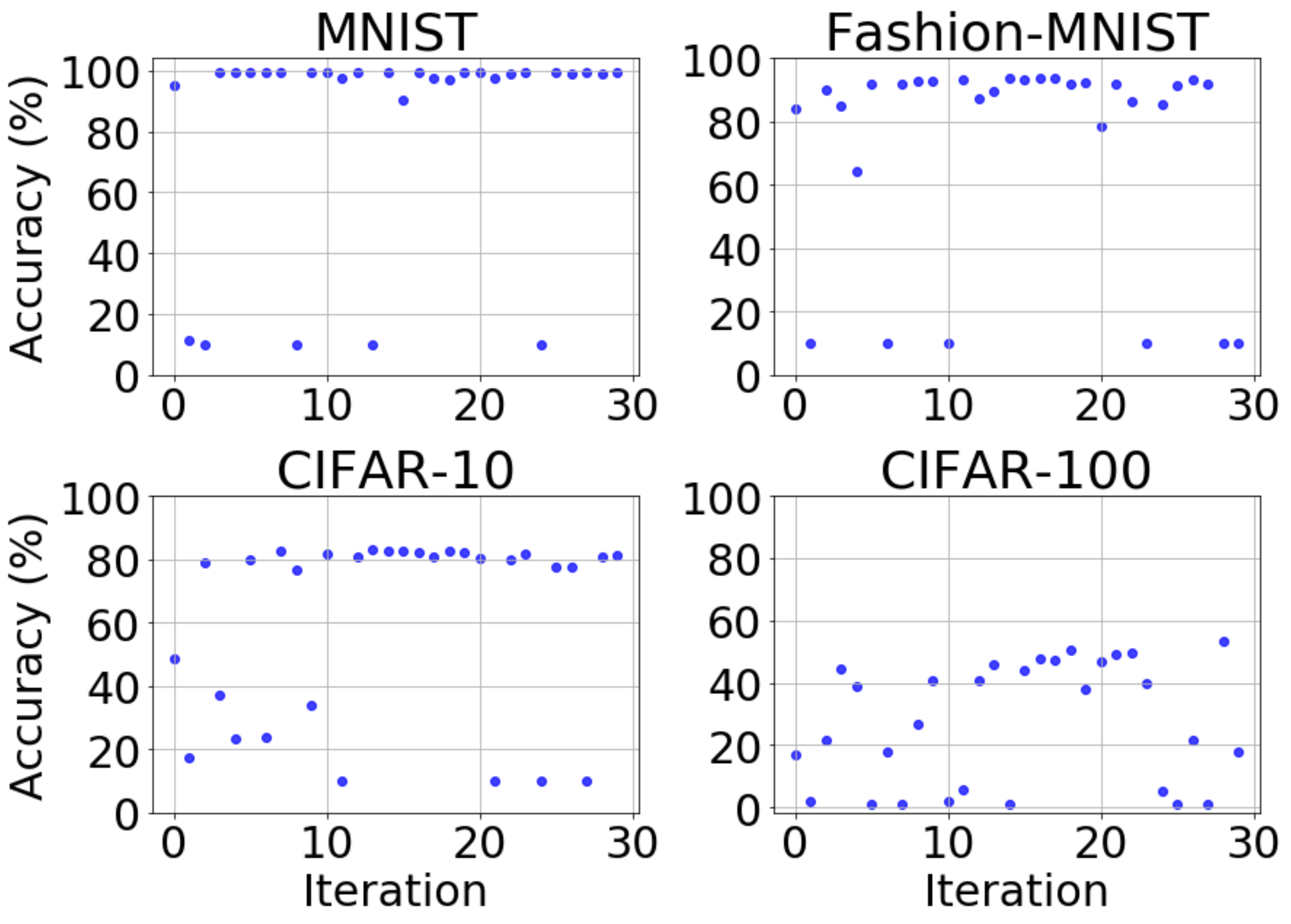}
    \caption{Case study two: Performance value (accuracy (\%) for each hyperparameter optimization search iteration for MNIST, Fashion-MNIST, CIFAR-10, and CIFAR-100 dataset with the hyperparameters given in Table~\ref{tab:cs2_hp}}
    \label{fig:cs2_acc}
\vspace{-0.2cm}
\end{figure}

\begin{figure}
    \centering
    \includegraphics [scale=.28]{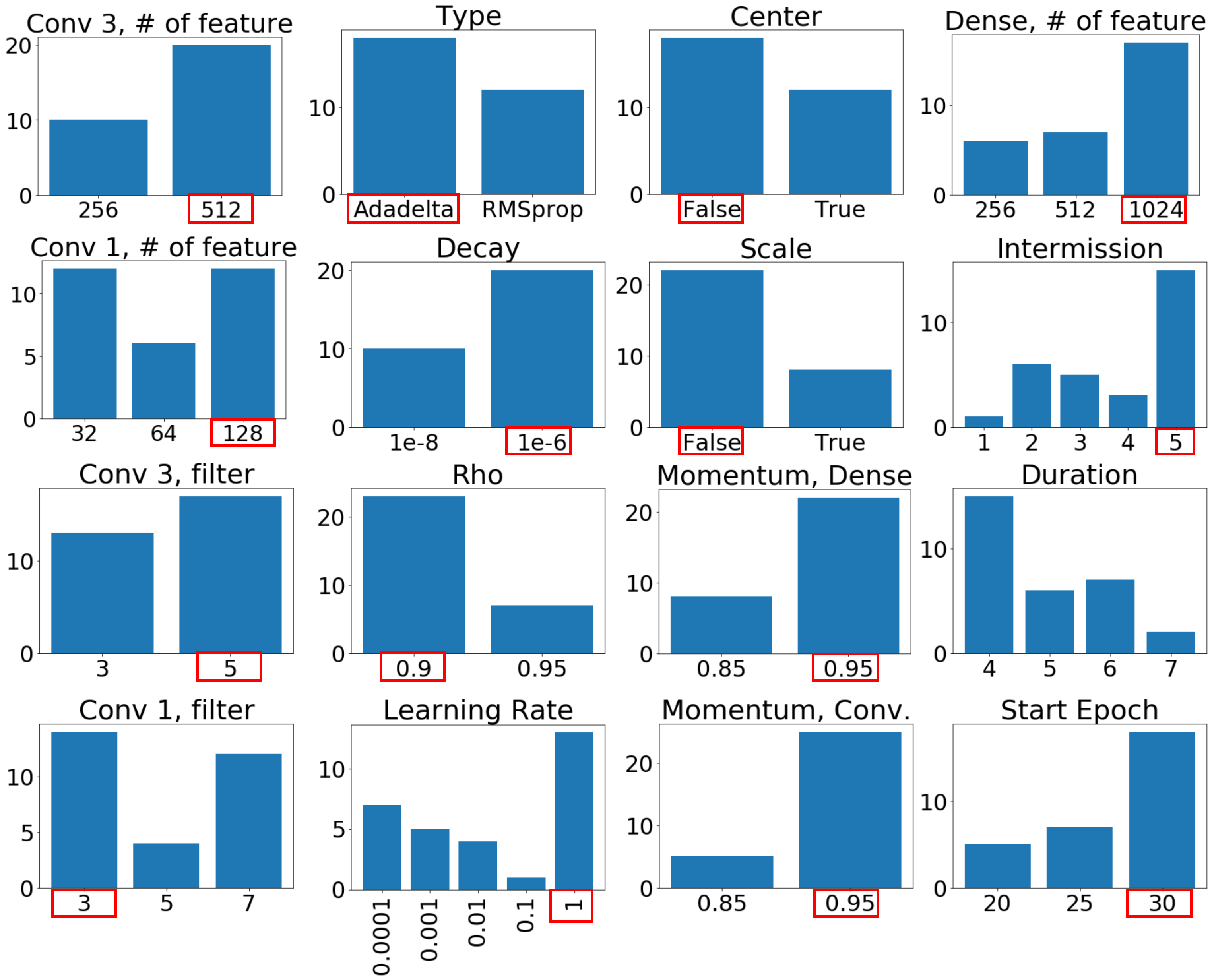}
    \caption{Case study two: Histograms of each hyperparameter value for CIFAR-100 dataset experiment for the 30 iterations of the optimization search process}
    \label{fig:cs2_hist}
\vspace{-0.5cm}    
\end{figure}

Figure~\ref{fig:cs2_acc} demonstrates the exploration and exploitation capability of Bayesian optimization technique in finding the optimum set of hyperparameter for each dataset. Starting from two random sets of hyperparameters, the search technique not only exploits and leverages the sets of hyperparameters with decent performance, but also explores the search space. In Figure~\ref{fig:cs2_hist}, we show the frequency of selecting each value for some of the hyperparameters given in Table~\ref{tab:cs2_hp} for CIFAR-100. The x-axis is the choice of hyperparameter and the y-axis is the number of times that a specific choice is called within the 30 evaluations in the Bayesian optimization search. The optimum hyperparameter values are highlighted in red rectangles in the figure. This also shows that after 30 iterations for searching the optimum hyperparameter set, the Bayesian framework not only leans toward the optimum values by selecting them most, but also tries all possible hyperparameter values to avoid trapping in any local minimum.

\begin{table*}[]\centering
\caption{Comparison of the SNN classification accuracies on MNIST, Fashion-MNIST, CIFAR-10, and CIFAR-100 dataset}
\vspace{-0.2cm}
\label{tab:literature}
\begin{tabular}{l|l|l|c|c|c|c}
\multirow{2}{*}{Model} & \multirow{2}{*}{Network Architecture} & \multirow{2}{*}{Method} & \multicolumn{4}{c}{Accuracy (\%)} \\ 
 &  &  & MNIST & \begin{tabular}[c]{@{}c@{}}Fashion\\ MNIST\end{tabular} & CIFAR-10 & CIFAR-100 \\ \hline \hline
Shrestha et al. \cite{shrestha2018slayer} & 6-layer CNN & \begin{tabular}[c]{@{}l@{}}Temporal credit assignment \\ for backpropagating (BP) error\end{tabular} & 99.36 & - & - & - \\
Rueckauer et al. \cite{rueckauer2017conversion} & 8-layer CNN & Offline, ANN-to-SNN conversion & 99.44 & - & 90.85 & - \\
Hunsberger et al. \cite{hunsberger2016training} & AlexNet & Offline, ANN-to-SNN conversion & 99.12 & - & 83.54 & 55.13 \\
Lee et al. \cite{lee2019enabling} & ResNet-11 & Spike-based backpropagating & 99.59 & - & 90.95 & - \\
Hao et al. \cite{hao2020biologically} & 3-layer FF SNN & Symmetric STDP Rule & 96.73 & 85.31 & - & - \\
Shrestha et al. \cite{shrestha2019approximating} & 4-layer NN & \begin{tabular}[c]{@{}l@{}}Error Modulated STDP\\  with symmetric weights\end{tabular} & 97.3 & 86.1 & - & - \\
Jin et al \cite{jin2018hybrid} & 6-layer CNN & Direct macro/micro BP & 99.49 & - & - & - \\
Sengupta et al. \cite{sengupta2019going} & VGG-16 & Offline, ANN-to-SNN conversion & - & - & 91.55 & - \\
Machado et al. \cite{machado2019natcsnn} & 3-layer NatCSNN & \begin{tabular}[c]{@{}l@{}}Two-phase (unsupervised STDP,\\ ReSuMe supervised)\end{tabular} & - & - & 84.7 & - \\
Wu et al. \cite{wutandem} & CIFARNet & SNN and ANN with shared weights & - & - & 91.54 & - \\
Roy et al. \cite{roy2019scaling} & VGG-9 & StochSigmoid XNOR-Net & - & - & 87.95 & 55.54 \\
Xing et al. \cite{xing2019homeostasis} & Inception-v4 & Homeostasis-based conversion & - & - & 92.49 & 70.4 \\
Hu et al. \cite{hu2018spiking} & ResNet-8 & ANN-to-SNN conversion & 99.59 & - & - & - \\
Hu et al. \cite{hu2018spiking} & ResNet-44 & ANN-to-SNN conversion & - & - & 91.98 & 68.56 \\
Guerguiev et al. \cite{guerguiev2019spike} & ConvNet + LIFNet & Regression discontinuity design & - & 91.81 & 76.2 & - \\
Thiele et al. \cite{thiele2019spikegrad} &  & Direct spike gradient & 99.52 & - & 89.99 & - \\
Wu et al. \cite{wu2019direct} & 8-layer CNN & Error BP through time & - & - & 90.53 & - \\
Severa et al. \cite{severa2019training} & VGG-like & Whetstone (Sharpened ANN) & 99.53 & - & 84.67 & - \\ \hline
Severa et al. \cite{severa2019training} & 6-layer CNN & Whetstone (Sharpened ANN) & 99.53 & 93.2 & 79 & 38 \\ 
\textbf{\begin{tabular}[c]{@{}l@{}}Hyperparameter Optimized\\ Whetstone (this work)\end{tabular}} & \textbf{6-layer CNN} & \textbf{\begin{tabular}[c]{@{}l@{}}Bayesian hyperparameter\\  optimized Whetstone\end{tabular}} & \textbf{99.6} & \textbf{93.68} & \textbf{84.36} & \textbf{53.42}
\end{tabular}
\end{table*}

Table~\ref{tab:sensitivity} gives a comprehensive sensitivity analysis on changing hyperparameter values and observing the final performance of the spiking neural network for CIFAR-100 dataset with the hyperparameter values given in Table~\ref{tab:cs2_hp}, and the performances shown in Figure~\ref{fig:cs2_acc} for this dataset. These experiments are chosen among the 30 iterations of the Bayesian optimization search. The first three experiments in Table~\ref{tab:sensitivity} show that with quite different combinations of hyperparameters we are getting almost zero improvement in the classification performance. This also intuitively shows that when the performance is not acceptable, the Bayesian approach drastically changes the hyperparameters to find the areas in the search space with better accuracies. In experiment four, the hyperparameter combination leads to an acceptable classification performance of $44.21\%$. From this point forward, the changes in the hyperparameter values are less aggressive to leverage the decent performance (only two hyperparameter values are changed from experiment four to five). In experiment six, optimizer hyperparameter type and the corresponding learning rate are changed; however, the final performance is within the same range compared to experiment five. This shows that different sets of hyperparameters might lead to similar classification performances. This indicates that this problem is well-suited for multi-objective hyperparameter optimization problems, where we might achieve similar performance while minimizing energy or area consumption. Experiments seven and eight demonstrate the exploration aspect of our optimization approach, meaning that although we already know an acceptable values for the hyperparameters, we also explore other areas of the search space to see if we can further improve the performance or not.

\begin{table*}[]
\caption{Case study two, Sensitivity Analysis: Comparing CIFAR-100 classification accuracy for different experiments}
\label{tab:sensitivity}
\begin{tabular}{l|c|c|c|c|c|c|c|c|c}
\multicolumn{1}{c|}{\textbf{CIFAR-100}} & \textbf{Exp. 1} & \textbf{Exp. 2} & \textbf{Exp.3} & \textbf{Exp. 4} & \textbf{Exp. 5} & \textbf{Exp. 6} & \textbf{Exp. 7} & \textbf{Exp. 8} & \textbf{Exp. 9} \\ \hline \hline
Optimizer Learning Rate & 0.0001 & 1 & 0.1 & 0.0001 & 0.0001 & 1 & 1 & 0.001 & 1 \\
Optimizer Rho & 0.9 & 0.9 & 0.9 & 0.9 & 0.9 & 0.9 & 0.9 & 0.9 & 0.9 \\
Optimizer Epsilon & 1e-8 & 1e-6 & 1e-8 & 1e-8 & 1e-8 & 1e-8 & 1e-8 & 1e-6 & 1e-6 \\
Optimizer Decay & 1e-8 & 1e-8 & 1e-6 & 1e-6 & 1e-6 & 1e-6 & 1e-6 & 1e-8 & 1e-6 \\
Optimizer Type & Adadelta & RMSprop & RMSprop & RMSprop & RMSProp & Adadelta & RMSprop & Adadelta & Adadelta \\
Noise Standard deviation & 0.2 & 0.3 & - & - & - & - & 0.3 & 0.3 & - \\
Noise Location & 1st Dense & 1st Dense & No Noise & No Noise & No Noise & No Noise & 1st Dense & 1st Dense & No Noise \\
Batch Norm. Momentum, conv. & 0.95 & 0.85 & 0.95 & 0.95 & 0.95 & 0.95 & 0.95 & 0.95 & 0.95 \\
Batch Norm. Momentum, dense & 0.85 & 0.95 & 0.95 & 0.85 & 0.95 & 0.95 & 0.95 & 0.95 & 0.95 \\
Batch Norm. Epsilon & 1e-3 & 1e-2 & 1e-2 & 1e-2 & 1e-2 & 1e-2 & 1e-3 & 1e-3 & 1e-3 \\
Batch Norm. Center & True & True & False & False & False & False & False & False & False \\
Batch Norm. Scale & True & True & True & True & False & False & False & False & False \\
Sharpener Start Epoch & 25 & 20 & 25 & 30 & 30 & 30 & 20 & 25 & 30 \\
Sharpener Duration & 7 & 6 & 4 & 4 & 4 & 4 & 4 & 4 & 4 \\
Sharpener Intermission & 2 & 1 & 5 & 5 & 5 & 5 & 5 & 5 & 5 \\
Conv. layer 1, filter size & 7 & 5 & 3 & 3 & 3 & 3 & 3 & 3 & 3 \\
Conv. layer 2, filter size & 5 & 5 & 5 & 5 & 5 & 5 & 5 & 5 & 5 \\
Conv. layer 3, filter size & 3 & 3 & 5 & 5 & 5 & 5 & 5 & 5 & 5 \\
Conv. layer 1, \# of features & 64 & 128 & 32 & 32 & 32 & 128 & 128 & 128 & 128 \\
Conv. layer 2, \# of features & 256 & 128 & 64 & 64 & 64 & 64 & 128 & 256 & 256 \\
Conv. layer 3, \# of features & 512 & 512 & 256 & 512 & 512 & 512 & 512 & 256 & 512 \\
Dense layer, \# of features & 256 & 1024 & 256 & 1024 & 1024 & 1024 & 1024 & 1024 & 1024 \\ \hline \hline
\textbf{Accuracy} & \textbf{1.96\%} & \textbf{1.01\%} & \textbf{1.01\%} & \textbf{44.21\%} & \textbf{46.07\%} & \textbf{48\%} & \textbf{1.01\%} & \textbf{21.73\%} & \textbf{53.42\%}
\end{tabular}
\end{table*}


\section{Discussion and Conclusion}

In this work, we introduce a hyperparameter optimization approach on Whetstone for training neural networks that can be deployed to neuromorphic hardware.  We show that by optimizing the hyperparameters associated with Whetstone we increase the performance over the previous state-of-the-art for this algorithm.  From our results, we see that the choice of hyperparameters (even among reasonable choices) can have a tremendous effect on the performance of Whetstone.  We also observe that the best hyperparameters found for each dataset differ across the datasets, indicating the importance of specifically optimizing hyperparameters for each new problem when converting to binary communication.  
We perform some small network architecture optimizations in this work.  In particular, we optimize the filter size and number of features for each of the three convolutional layers, as well as the number of features for the dense layer. We are limiting our search to a fixed maximum network depth to deploy it on embedded systems in the future. The best results on the different datasets are shown with different parameters in Table \ref{tab:cs2_acc}.  We anticipate that further optimizing the network architecture will be able to improve the performance of Whetstone on different datasets. In future work, we plan to use an optimization approach such as MENNDL \cite{young2017evolving} to further optimize the architecture (the number and type of layers) of these networks.  Whetstone's simple modifications to neural network design should allow us to search for topologies including sharpening activations within the MENNDL framework to better understand when sharpening is useful and hopefully discover higher performance network designs that may better leverage binarized operations.

In~\cite{vineyard2019low}, Whetstone is deployed on SpiNNaker \cite{furber2014spinnaker}, with slight drop in accuracy due to issues with input/output encoding. Here, we optimize the network using Whetstone, but we do not map the resulting networks to a neuromorphic hardware implementation, such as SpiNNaker \cite{furber2014spinnaker} or Loihi \cite{davies2018loihi}. As observed in~\cite{vineyard2019low}, several other hyperparameters such as input/output encoding, different network topologies and training parameters will have an effect on this mapping performance.  In the future, we plan to include how the network performs on real neuromorphic hardware as part of our training objectives in the hyperparameter and network architecture optimization process.

Finally, as we consider mapping onto real neuromorphic hardware, there are often other important performance considerations beyond accuracy on the task at hand.  For example, size, area, and energy efficiency are often important considerations for real deployments of neuromorphic systems. As such, it is important to train with those objectives in mind.  In previous work, we have extended the Bayesian optimization approach \cite{parsa2020journal, parsa2019pabo} and the fitness function used within MENNDL \cite{young2019evolving} to incorporate multiple objectives. In future work, we plan to apply this approach to the Whetstone algorithm in order to optimize networks that are both more accurate, but also more efficient.

    

\section*{Acknowledgment}

The research was funded in part by Center for Brain-Inspired Computing Enabling Autonomous Intelligence (C-BRIC), one of six centers in JUMP, a Semiconductor Research Corporation (SRC) program sponsored by DARPA, the National Science Foundation, Intel Corporation and Vannevar Bush Faculty Fellowship.

This material is also based in part upon work supported by the U.S. Department of Energy, Office of Science, Office of Advanced Scientific Computing Research, under contract number DE-AC05-00OR22725, and in part by the Laboratory Directed Research and Development Program of Oak Ridge National Laboratory, managed by UT-Battelle, LLC.

Sandia National Laboratories is a multimission laboratory managed and operated by National Technology \& Engineering Solutions of Sandia, LLC, a wholly owned subsidiary of Honeywell International Inc., for the U.S. Department of Energy's National Nuclear Security Administration under contract DE-NA0003525.

\bibliographystyle{IEEEtran}
\bibliography{MP}

\begin{thebibliography}{10}
\providecommand{\url}[1]{#1}
\csname url@samestyle\endcsname
\providecommand{\newblock}{\relax}
\providecommand{\bibinfo}[2]{#2}
\providecommand{\BIBentrySTDinterwordspacing}{\spaceskip=0pt\relax}
\providecommand{\BIBentryALTinterwordstretchfactor}{4}
\providecommand{\BIBentryALTinterwordspacing}{\spaceskip=\fontdimen2\font plus
\BIBentryALTinterwordstretchfactor\fontdimen3\font minus
  \fontdimen4\font\relax}
\providecommand{\BIBforeignlanguage}[2]{{%
\expandafter\ifx\csname l@#1\endcsname\relax
\typeout{** WARNING: IEEEtran.bst: No hyphenation pattern has been}%
\typeout{** loaded for the language `#1'. Using the pattern for}%
\typeout{** the default language instead.}%
\else
\language=\csname l@#1\endcsname
\fi
#2}}
\providecommand{\BIBdecl}{\relax}
\BIBdecl

\bibitem{aimone2018non}
J.~B. Aimone, K.~E. Hamilton, S.~Mniszewski, L.~Reeder, C.~D. Schuman, and
  W.~M. Severa, ``Non-neural network applications for spiking neuromorphic
  hardware,'' in \emph{3rd International Workshop on Post-Moore's Era
  Supercomputing (PMES 2018), Dallas, TX}, 2018.

\bibitem{schuman2017survey}
C.~D. Schuman, T.~E. Potok, R.~M. Patton, J.~D. Birdwell, M.~E. Dean, G.~S.
  Rose, and J.~S. Plank, ``A survey of neuromorphic computing and neural
  networks in hardware,'' \emph{arXiv preprint arXiv:1705.06963}, 2017.

\bibitem{bergstra2013hyperopt}
J.~Bergstra, D.~Yamins, and D.~D. Cox, ``Hyperopt: A python library for
  optimizing the hyperparameters of machine learning algorithms,'' in
  \emph{Proceedings of the 12th Python in science conference}.\hskip 1em plus
  0.5em minus 0.4em\relax Citeseer, 2013, pp. 13--20.

\bibitem{hernandez2016general}
J.~M. Hern{\'a}ndez-Lobato, M.~A. Gelbart, R.~P. Adams, M.~W. Hoffman, and
  Z.~Ghahramani, ``A general framework for constrained bayesian optimization
  using information-based search,'' \emph{The Journal of Machine Learning
  Research}, vol.~17, no.~1, pp. 5549--5601, 2016.

\bibitem{parsa2019pabo}
M.~Parsa, A.~Ankit, A.~Ziabari, and K.~Roy, ``Pabo: Pseudo agent-based
  multi-objective bayesian hyperparameter optimization for efficient neural
  accelerator design,'' in \emph{2019 IEEE/ACM International Conference on
  Computer-Aided Design (ICCAD)}, 2019, pp. 1--8.

\bibitem{bergstra2012random}
J.~Bergstra and Y.~Bengio, ``Random search for hyper-parameter optimization,''
  \emph{Journal of Machine Learning Research}, vol.~13, no. Feb, pp. 281--305,
  2012.

\bibitem{severa2019training}
W.~Severa, C.~M. Vineyard, R.~Dellana, S.~J. Verzi, and J.~B. Aimone,
  ``Training deep neural networks for binary communication with the whetstone
  method,'' \emph{Nature Machine Intelligence}, vol.~1, no.~2, p.~86, 2019.

\bibitem{parsa2019bayesian}
M.~Parsa, J.~P. Mitchell, C.~D. Schuman, R.~M. Patton, T.~E. Potok, and K.~Roy,
  ``Bayesian-based hyperparameter optimization for spiking neuromorphic
  systems,'' in \emph{2019 IEEE International Conference on Big Data (Big
  Data)}.\hskip 1em plus 0.5em minus 0.4em\relax IEEE, 2019, pp. 4472--4478.

\bibitem{parsa2020journal}
------, ``Bayesian multi-objective hyperparameter optimization for accurate,
  fast, and efficient neural network accelerator design,'' 2020, p. submitted.

\bibitem{date2016design}
P.~Date, J.~A. Hendler, and C.~D. Carothers, ``Design index for deep neural
  networks,'' \emph{Procedia Computer Science}, vol.~88, pp. 131--138, 2016.

\bibitem{bengio2012practical}
Y.~Bengio, ``Practical recommendations for gradient-based training of deep
  architectures,'' in \emph{Neural networks: Tricks of the trade}.\hskip 1em
  plus 0.5em minus 0.4em\relax Springer, 2012, pp. 437--478.

\bibitem{bergstra2011algorithms}
J.~S. Bergstra, R.~Bardenet, Y.~Bengio, and B.~K{\'e}gl, ``Algorithms for
  hyper-parameter optimization,'' in \emph{Advances in neural information
  processing systems}, 2011, pp. 2546--2554.

\bibitem{bergstra2014preliminary}
J.~Bergstra, B.~Komer, C.~Eliasmith, and D.~Warde-Farley, ``Preliminary
  evaluation of hyperopt algorithms on hpolib,'' in \emph{ICML workshop on
  AutoML}, 2014.

\bibitem{snoek2012practical}
J.~Snoek, H.~Larochelle, and R.~P. Adams, ``Practical bayesian optimization of
  machine learning algorithms,'' in \emph{Advances in neural information
  processing systems}, 2012, pp. 2951--2959.

\bibitem{zhang2015improving}
Y.~Zhang, K.~Sohn, R.~Villegas, G.~Pan, and H.~Lee, ``Improving object
  detection with deep convolutional networks via bayesian optimization and
  structured prediction,'' in \emph{Proceedings of the IEEE Conference on
  Computer Vision and Pattern Recognition}, 2015, pp. 249--258.

\bibitem{balaprakash2018deephyper}
P.~Balaprakash, M.~Salim, T.~Uram, V.~Vishwanath, and S.~Wild, ``Deephyper:
  Asynchronous hyperparameter search for deep neural networks,'' in \emph{2018
  IEEE 25th International Conference on High Performance Computing
  (HiPC)}.\hskip 1em plus 0.5em minus 0.4em\relax IEEE, 2018, pp. 42--51.

\bibitem{ilievski2017efficient}
I.~Ilievski, T.~Akhtar, J.~Feng, and C.~A. Shoemaker, ``Efficient
  hyperparameter optimization for deep learning algorithms using deterministic
  rbf surrogates,'' in \emph{Thirty-First AAAI Conference on Artificial
  Intelligence}, 2017.

\bibitem{miikkulainen2019evolving}
R.~Miikkulainen, J.~Liang, E.~Meyerson, A.~Rawal, D.~Fink, O.~Francon, B.~Raju,
  H.~Shahrzad, A.~Navruzyan, N.~Duffy \emph{et~al.}, ``Evolving deep neural
  networks,'' in \emph{Artificial Intelligence in the Age of Neural Networks
  and Brain Computing}.\hskip 1em plus 0.5em minus 0.4em\relax Elsevier, 2019,
  pp. 293--312.

\bibitem{young2017evolving}
S.~R. Young, D.~C. Rose, T.~Johnston, W.~T. Heller, T.~P. Karnowski, T.~E.
  Potok, R.~M. Patton, G.~Perdue, and J.~Miller, ``Evolving deep networks using
  hpc,'' in \emph{Proceedings of the Machine Learning on HPC
  Environments}.\hskip 1em plus 0.5em minus 0.4em\relax ACM, 2017, p.~7.

\bibitem{shafiee2018deep}
M.~J. Shafiee, A.~Mishra, and A.~Wong, ``Deep learning with darwin:
  Evolutionary synthesis of deep neural networks,'' \emph{Neural Processing
  Letters}, vol.~48, no.~1, pp. 603--613, 2018.

\bibitem{liang2018evolutionary}
J.~Liang, E.~Meyerson, and R.~Miikkulainen, ``Evolutionary architecture search
  for deep multitask networks,'' in \emph{Proceedings of the Genetic and
  Evolutionary Computation Conference}.\hskip 1em plus 0.5em minus 0.4em\relax
  ACM, 2018, pp. 466--473.

\bibitem{schuman2019non}
C.~D. Schuman, J.~S. Plank, G.~Bruer, and J.~Anantharaj, ``Non-traditional
  input encoding schemes for spiking neuromorphic systems,'' in \emph{2019
  International Joint Conference on Neural Networks (IJCNN)}.\hskip 1em plus
  0.5em minus 0.4em\relax IEEE, 2019, pp. 1--10.

\bibitem{salt2017differential}
L.~Salt, D.~Howard, G.~Indiveri, and Y.~Sandamirskaya, ``Differential evolution
  and bayesian optimisation for hyper-parameter selection in mixed-signal
  neuromorphic circuits applied to uav obstacle avoidance,'' \emph{arXiv
  preprint arXiv:1704.04853}, 2017.

\bibitem{schuman2016evolutionary}
C.~D. Schuman, J.~S. Plank, A.~Disney, and J.~Reynolds, ``An evolutionary
  optimization framework for neural networks and neuromorphic architectures,''
  in \emph{2016 International Joint Conference on Neural Networks
  (IJCNN)}.\hskip 1em plus 0.5em minus 0.4em\relax IEEE, 2016, pp. 145--154.

\bibitem{kim2018competitive}
J.~Kim and D.-S. Kim, ``Competitive hyperparameter balancing on spiking neural
  network for a fast, accurate and energy-efficient inference,'' in
  \emph{International Symposium on Neural Networks}.\hskip 1em plus 0.5em minus
  0.4em\relax Springer, 2018, pp. 44--53.

\bibitem{shahriari2015taking}
B.~Shahriari, K.~Swersky, Z.~Wang, R.~P. Adams, and N.~De~Freitas, ``Taking the
  human out of the loop: A review of bayesian optimization,'' \emph{Proceedings
  of the IEEE}, vol. 104, no.~1, pp. 148--175, 2015.

\bibitem{skopt:2020}
\BIBentryALTinterwordspacing
\emph{Scikit-Optimize python package}, accessed January 6, 2020. [Online].
  Available: \url{https://scikit-optimize.github.io/}
\BIBentrySTDinterwordspacing

\bibitem{lecun2010mnist}
Y.~LeCun, C.~Cortes, and C.~Burges, ``Mnist handwritten digit database,''
  \emph{ATT Labs [Online]. Available: http://yann. lecun. com/exdb/mnist},
  vol.~2, 2010.

\bibitem{xiao2017fashion}
H.~Xiao, K.~Rasul, and R.~Vollgraf, ``Fashion-mnist: a novel image dataset for
  benchmarking machine learning algorithms,'' \emph{arXiv preprint
  arXiv:1708.07747}, 2017.

\bibitem{krizhevsky2009learning}
A.~Krizhevsky, G.~Hinton \emph{et~al.}, ``Learning multiple layers of features
  from tiny images,'' 2009.

\bibitem{shrestha2018slayer}
S.~B. Shrestha and G.~Orchard, ``Slayer: Spike layer error reassignment in
  time,'' in \emph{Advances in Neural Information Processing Systems}, 2018,
  pp. 1412--1421.

\bibitem{rueckauer2017conversion}
B.~Rueckauer, I.-A. Lungu, Y.~Hu, M.~Pfeiffer, and S.-C. Liu, ``Conversion of
  continuous-valued deep networks to efficient event-driven networks for image
  classification,'' \emph{Frontiers in neuroscience}, vol.~11, p. 682, 2017.

\bibitem{hunsberger2016training}
E.~Hunsberger and C.~Eliasmith, ``Training spiking deep networks for
  neuromorphic hardware,'' \emph{arXiv preprint arXiv:1611.05141}, 2016.

\bibitem{lee2019enabling}
C.~Lee, S.~S. Sarwar, and K.~Roy, ``Enabling spike-based backpropagation in
  state-of-the-art deep neural network architectures,'' \emph{arXiv preprint
  arXiv:1903.06379}, 2019.

\bibitem{hao2020biologically}
Y.~Hao, X.~Huang, M.~Dong, and B.~Xu, ``A biologically plausible supervised
  learning method for spiking neural networks using the symmetric stdp rule,''
  \emph{Neural Networks}, vol. 121, pp. 387--395, 2020.

\bibitem{shrestha2019approximating}
A.~Shrestha, H.~Fang, Q.~Wu, and Q.~Qiu, ``Approximating back-propagation for a
  biologically plausible local learning rule in spiking neural networks,'' in
  \emph{Proceedings of the International Conference on Neuromorphic Systems},
  2019, pp. 1--8.

\bibitem{jin2018hybrid}
Y.~Jin, W.~Zhang, and P.~Li, ``Hybrid macro/micro level backpropagation for
  training deep spiking neural networks,'' in \emph{Advances in Neural
  Information Processing Systems}, 2018, pp. 7005--7015.

\bibitem{sengupta2019going}
A.~Sengupta, Y.~Ye, R.~Wang, C.~Liu, and K.~Roy, ``Going deeper in spiking
  neural networks: Vgg and residual architectures,'' \emph{Frontiers in
  neuroscience}, vol.~13, 2019.

\bibitem{machado2019natcsnn}
P.~Machado, G.~Cosma, and T.~M. McGinnity, ``Natcsnn: A convolutional spiking
  neural network for recognition of objects extracted from natural images,'' in
  \emph{International Conference on Artificial Neural Networks}.\hskip 1em plus
  0.5em minus 0.4em\relax Springer, 2019, pp. 351--362.

\bibitem{wutandem}
J.~Wu, Y.~Chua, M.~Zhang, G.~Li, H.~Li, and K.~C. Tan, ``A tandem learning rule
  for efficient and rapid inference on deep spiking neural networks.''

\bibitem{roy2019scaling}
D.~Roy, I.~Chakraborty, and K.~Roy, ``Scaling deep spiking neural networks with
  binary stochastic activations,'' in \emph{2019 IEEE International Conference
  on Cognitive Computing (ICCC)}.\hskip 1em plus 0.5em minus 0.4em\relax IEEE,
  2019, pp. 50--58.

\bibitem{xing2019homeostasis}
F.~Xing, Y.~Yuan, H.~Huo, and T.~Fang, ``Homeostasis-based cnn-to-snn
  conversion of inception and residual architectures,'' in \emph{International
  Conference on Neural Information Processing}.\hskip 1em plus 0.5em minus
  0.4em\relax Springer, 2019, pp. 173--184.

\bibitem{hu2018spiking}
Y.~Hu, H.~Tang, Y.~Wang, and G.~Pan, ``Spiking deep residual network,''
  \emph{arXiv preprint arXiv:1805.01352}, 2018.

\bibitem{guerguiev2019spike}
J.~Guerguiev, K.~P. Kording, and B.~A. Richards, ``Spike-based causal inference
  for weight alignment,'' \emph{arXiv preprint arXiv:1910.01689}, 2019.

\bibitem{thiele2019spikegrad}
J.~C. Thiele, O.~Bichler, and A.~Dupret, ``Spikegrad: An ann-equivalent
  computation model for implementing backpropagation with spikes,'' \emph{arXiv
  preprint arXiv:1906.00851}, 2019.

\bibitem{wu2019direct}
Y.~Wu, L.~Deng, G.~Li, J.~Zhu, Y.~Xie, and L.~Shi, ``Direct training for
  spiking neural networks: Faster, larger, better,'' in \emph{Proceedings of
  the AAAI Conference on Artificial Intelligence}, vol.~33, 2019, pp.
  1311--1318.

\bibitem{vineyard2019low}
C.~M. Vineyard, R.~Dellana, J.~B. Aimone, F.~Rothganger, and W.~M. Severa,
  ``Low-power deep learning inference using the spinnaker neuromorphic
  platform,'' in \emph{Proceedings of the 7th Annual Neuro-inspired
  Computational Elements Workshop}, 2019, pp. 1--7.

\bibitem{furber2014spinnaker}
S.~B. Furber, F.~Galluppi, S.~Temple, and L.~A. Plana, ``The spinnaker
  project,'' \emph{Proceedings of the IEEE}, vol. 102, no.~5, pp. 652--665,
  2014.

\bibitem{davies2018loihi}
M.~Davies, N.~Srinivasa, T.-H. Lin, G.~Chinya, Y.~Cao, S.~H. Choday, G.~Dimou,
  P.~Joshi, N.~Imam, S.~Jain \emph{et~al.}, ``Loihi: A neuromorphic manycore
  processor with on-chip learning,'' \emph{IEEE Micro}, vol.~38, no.~1, pp.
  82--99, 2018.

\bibitem{young2019evolving}
S.~R. Young, P.~Devineni, M.~Parsa, J.~T. Johnston, B.~Kay, R.~M. Patton, C.~D.
  Schuman, D.~C. Rose, and T.~E. Potok, ``Evolving energy efficient
  convolutional neural networks,'' in \emph{2019 IEEE International Conference
  on Big Data (Big Data)}.\hskip 1em plus 0.5em minus 0.4em\relax IEEE, 2019,
  pp. 4479--4485.

\end{thebibliography}

\end{document}